\documentclass[sigconf,screen]{acmart} 
\pdfoutput=1

\usepackage[utf8]{inputenc}
\usepackage[T1]{fontenc}
\usepackage{microtype}

\usepackage{tcolorbox}
\tcbuselibrary{most}

\usepackage{epstopdf}
\usepackage{bbm}
\usepackage{dsfont}

\usepackage{listings}

\usepackage{algorithm}
\usepackage{algorithmic}

\usepackage{amsmath}

\usepackage{multirow}
\usepackage{graphicx}
\usepackage{graphics}
\usepackage{lscape}
\usepackage{bm}
\usepackage{setspace}
\usepackage{verbatim}
\usepackage{todonotes}
\usepackage{url}
\usepackage{hyperref}
\usepackage{colortbl}
\usepackage{enumitem}
\usepackage{booktabs}
\usepackage{balance}

\usepackage{natbib}
\usepackage{doi}

\hypersetup
{
colorlinks=true,
linkcolor=blue,
filecolor=blue,
urlcolor=blue,
citecolor=cyan,
}

\usepackage[labelfont=bf]{caption}

\AtBeginDocument{%
  \providecommand\BibTeX{{%
    \normalfont B\kern-0.5em{\scshape i\kern-0.25em b}\kern-0.8em\TeX}}}
    
\setcopyright{acmcopyright}
\acmPrice{15.00}
\acmDOI{10.1145/3540250.3549103}
\acmYear{2022}
\copyrightyear{2022}
\acmSubmissionID{fse22main-p210-p}
\acmISBN{978-1-4503-9413-0/22/11}
\acmConference[ESEC/FSE '22]{Proceedings of the 30th ACM Joint European Software Engineering Conference and Symposium on the Foundations of Software Engineering}{November 14--18, 2022}{Singapore, Singapore}
\acmBooktitle{Proceedings of the 30th ACM Joint European Software Engineering Conference and Symposium on the Foundations of Software Engineering (ESEC/FSE '22), November 14--18, 2022, Singapore, Singapore}

\begin{document}

\title{Adaptive Fairness Improvement Based on Causality Analysis}

\author{Mengdi Zhang}
\email{mdzhang.2019@phdcs.smu.edu.sg}
\orcid{0000-0002-3239-4804}
\affiliation{%
  \institution{Singapore Management University}
  \country{Singapore}
}

\author{Jun Sun}
\email{junsun@smu.edu.sg}
\orcid{0000-0002-3545-1392}
\affiliation{%
  \institution{Singapore Management University}
  \country{Singapore}
}

\begin{abstract}
Given a discriminating neural network, the problem of fairness improvement is to systematically reduce discrimination without significantly scarifies its performance (i.e., accuracy). Multiple categories of fairness improving methods have been proposed for neural networks, including pre-processing, in-processing and post-processing. Our empirical study however shows that these methods are not always effective (e.g., they may improve fairness by paying the price of huge accuracy drop) or even not helpful (e.g., they may even worsen both fairness and accuracy). In this work, we propose an approach which adaptively chooses the fairness improving method based on causality analysis. That is, we choose the method based on how the neurons and attributes responsible for unfairness are distributed among the input attributes and the hidden neurons. Our experimental evaluation shows that our approach is effective (i.e., always identify the best fairness improving method) and efficient (i.e., with an average time overhead of 5 minutes).  
\end{abstract}



\maketitle

\section{Introduction}\label{sec:intro}
Neural networks have found their way into a variety of systems, including many which potentially have significant societal impact, such as personal credit rating~\cite{eggermont2004genetic}, criminal sentencing~\cite{ruoss2020learning, compas2016data}, face recognition~\cite{vstruc2009gabor} and resume shortlisting~\cite{roy2020machine}. While these neural networks often have high accuracy in these classification tasks, some concerning fairness issues have been observed as well~\cite{ruoss2020learning, buolamwini2018gender, chakraborty2019software, friedler2019comparative, biswas2020machine}. That is, the predictions made by these neural networks may be biased with regard to certain protected attributes such as sex, race, and gender. For instance, it has been shown~\cite{bellamy2019ai} that a neural network trained to predict people's income level based on an individual’s personal information (which can be used in applications such as bank loan approval) is much more likely to predict male individuals with high-income level. Further analysis shows that for many individuals, changing only the gender or race causes the output of the predictions to flip~\cite{galhotra2017fairness}. For another instance, it has been shown~\cite{compas2016data} that a machine learning model used to predict the recidivism risk for suspected criminals is more likely to mislabel black defendants as having a high recidivism risk.

In recent years, many methods and tools have been proposed to detect discrimination in neural networks systematically (e.g., through the so-called fairness testing~\cite{galhotra2017fairness, zhang2020white, angell2018themis, udeshi2018automated, ma2020metamorphic}), and more relevantly, to improve the fairness of neural networks~\cite{feldman2015certifying, kamiran2012data, zhang2018mitigating, kamishima2012fairness, celis2019classification, agarwal2018reductions, hardt2016equality, kamiran2012decision, pleiss2017fairness}. In general, existing fairness improving methods can be classified into three categories according to when the method is applied, e.g., pre-processing, in-processing and post-processing methods. Pre-processing methods~\cite{kamiran2012data, feldman2015certifying, calmon2017optimized, zemel2013learning} aim to reduce the bias in the training data so as to reduce the bias of model predictions; in-processing methods~\cite{celis2019classification, zhang2018mitigating, kamishima2012fairness, agarwal2018reductions, agarwal2019fair, kearns2018preventing} focus on the model and the training process; and post-processing methods~\cite{hardt2016equality, kamiran2012decision, pleiss2017fairness} modify the prediction results directly rather than the training data or the model. 

However, fairness improving is a complicated task and it is not always clear which method should be applied. As shown in Section~\ref{sec:empirical}, different fairness improving methods perform significantly differently on different models (which is consistent with the partial results reported in~\cite{biswas2020machine,chakraborty2019software,friedler2019comparative}). More importantly, applying the `wrong' method would not only lead to a huge loss in accuracy (e.g., the accuracy of the model trained on the COMPAS dataset drops by 35\% after applying the Reject Option post-processing method), but also lead to worsened fairness. For instance, out of 90 cases (i.e., combinations of model, protected attribute and fairness improving method) that we examined in Section~\ref{sec:empirical}, 20\% of them result in worsened fairness. Furthermore, a fairness improving method may be effective with respect to one protected attribute whilst being harmful with respect to another protected attribute. For instance, the fairness of the model trained on Adult Income dataset improves by around 4\% with respect to the \emph{gender} attribute after applying the Equalized Odds post-processing method and worsens by 20\% with respect to the \emph{race} attribute. Given that many of the fairness improving methods require significant effort and computing resource, it is infeasible to try all of them and identify the best performing one. It is thus important to have a systematic way of identifying the `right' method efficiently. 

In this work, we propose to choose the `right' fairness improving method based on causality analysis. Intuitively, the idea is to conduct causality analysis so as to understand the causes of the discrimination, i.e., whether a certain number of input attributes or hidden neurons are highly responsible for the unfairness. Formally, we use the probability of high causal effects and Coefficient of Variation to characterize the distribution of the causal effects. Depending on the result of the causality analysis, we then choose the fairness improving method accordingly. For instance, if a small number of input attributes bare most of the responsibility for unfairness, a pre-processing method such as~\cite{kamiran2012data, feldman2015certifying} would be the choice, whereas an in-processing method would be the choice if some neurons are highly responsible. Our approach is designed based on the results of an empirical study which evaluates 9 fairness improving methods (i.e., 2 pre-processing methods, 4 in-processing methods and 3 post-processing methods) on 4 different benchmark datasets with respect to different fairness metrics. 
Our approach is systematically evaluated with the same models. The results show that our selected processing approach is the optimal choice to improve group fairness in all cases and the optimal choice to reduce individual discrimination in most cases. 

The remainders of the paper are organized as follows. In Section~\ref{sec:definition}, we review relevant background. In Section~\ref{sec:empirical}, we present results from our empirical study which motivates our approach. In Section~\ref{sec:adaptive_approach}, we present our adaptive fairness improving method. In Section~\ref{sec:evaluation}, we evaluate our approach. Lastly, we review related work in Section~\ref{sec:related} and conclude in Section~\ref{sec:conclusion}. 

\section{Background} \label{sec:definition}
In the following, we review relevant background on fairness and existing fairness improving methods. 

\subsection{Fairness Definitions}\label{sec:fairness improve}
In the literature, there are multiple definitions of fairness~\cite{dwork2012fairness,joseph2016fairness,calders2010three,galhotra2017fairness,kleinberg2016inherent,zafar2017fairness}. What is common across different definitions is that to define fairness, we must first identify a set of protected attributes (a.k.a.~sensitive attributes). Commonly recognized protected attributes instance race, sex, age and religion. Note that different models may have different protected attributes.

In the following, we introduce two popular definitions of fairness, i.e., group fairness and individual discrimination, as well as the corresponding fairness scores, i.e., metrics that are used to quantify the degree of unfairness. \\ 

\noindent \emph{Group fairness}, also known as statistical fairness, focuses on certain protected groups such as ethnic minority and the parity across different groups based on some statistical measurements. It is the primary focus of this study as well as many existing studies~\cite{zhang2021ignorance, harrison2020empirical, kearns2019empirical, causality2022, bellamy2019ai}. Classic measurements for group fairness include positive classification rate and true positive rate. A classifier satisfies group fairness if the samples in the protected groups have an equal or similar positive classification probability or true positive probability. 

Given a model, we can measure its degree of unfairness according to group fairness using Statistical Parity Difference (SPD)~\cite{calders2010three}\footnote{There are also alternative similar measures such as Disparate Impact~\cite{zafar2017fairness} that we omit in this study.}. 

\begin{definition}[Statistical Parity Difference]\label{def:sta_diff}
Let $Y$ be the predicted output of the neural network $N$; $l$ be a (favorable) prediction and $F$ be a protected attribute. 
Statistical Parity Difference is the difference in the probability of favorable outcomes between the unprivileged and privileged groups where the unprivileged/privileged groups are defined based on the value of the protected attribute. 
\begin{equation}
\begin{aligned}
|P[Y=l \mid F=0]-P[Y=l \mid F=1]|
\end{aligned}
\end{equation}
\end{definition}



Note that the above definition only considers a single binary protected attribute, which is sometimes insufficient. The following metric, called Group Discrimination Score (GDS), extends SPD to measure fairness based on multiple protected attributes.

\begin{definition}[Group Discrimination Score]\label{def:multi_group}
Let $N$ be a neural network; $Y$ be the predicted output of the neural network; $l$ be a (favorable) prediction, and $F$ be a set of (one or more) protected attributes. Let $\theta$ (and $\theta'$) be an arbitrary valuation of the protected attributes $F$. Let $X_\theta$ be the set of inputs whose $F$-attribute values are $\theta$. Let $P_\theta$ be $P(N(x) = l ~|~ x \in X_\theta)$. 
The multivariate group discrimination with respect to protected attributes $F$ is the maximum difference between any $P_\theta$ and $P_{\theta'}$.
\end{definition} 


\noindent \emph{Example} Consider the structured dataset Adult Income~\cite{census1996dataset}. It has two protected attributes, i.e., gender, and race. Each attribute has a set of two values, i.e., Female or Male for gender, and White or non-White for race. As a result, there are 4 possible $\theta$, i.e., (Male, White), (Female, White), (Male, non-White) and (Female, non-White). The probabilities of an individual who is predicted to have a high-income level (i.e., more than 50K) with respect to these four $\theta$ is 14.4\%, 39.6\%, 9.0\% and 28.5\% respectively. The GDS of the model is thus 30.6\%. \\

\noindent \emph{Individual discrimination} is another concept which is often applied in fairness analysis. It focuses on specific pairs of individuals. Intuitively, individual discrimination occurs when two individuals that differ by only certain protected attribute(s) are predicted with different labels. An individual whose label changes once its protected attribute(s) changes is referred to as an individual discriminatory instance. 

\begin{definition}[Individual Discriminatory Instance] \label{def:indi_dis_instance}
Let $F$ be a set of (one or more) protected attributes; and $N$ be a neural network. $x$ is an individual discriminatory instance if there exists an instance $x^{\prime}$ such that the following conditions are satisfied.
\begin{itemize}
\item $\forall q \not \in F.~x_q = x^{\prime}_q$
\item $N(x) \neq N(x^{\prime})$
\end{itemize}
\end{definition}
The above definition is often adopted in fairness testing, i.e., works on searching or generating individual discriminatory instances~\cite{zhang2020white,udeshi2018automated}. In addition, there are proposals on learning models which are more likely to avoid individual discriminatory~\cite{ruoss2020learning}. 

Given a model, we can measure its fairness according to individual discrimination by measuring the percentage of individual discriminatory instances in a set of instances (which can be the test set or a set generated to simulate unseen samples), formally called Causal Discrimination Score (CDS). 

\begin{definition}[Causal Discrimination Score]\label{def:uni_causal}
Let $N$ be a neural network; $F$ be a set of protected attributes. The causal discrimination score of $N$ with respect to protected attributes $F$, is the fraction of inputs which are individual discrimination instances. 
\end{definition}

\subsection{Fairness Improving Methods}
Many methods have been proposed to improve the fairness of neural networks~\cite{kamiran2012data, feldman2015certifying, calmon2017optimized, zemel2013learning, celis2019classification, zhang2018mitigating, kamishima2012fairness, agarwal2018reductions, agarwal2019fair, kearns2018preventing, hardt2016equality, pleiss2017fairness, kamiran2012decision}. They can be categorized into three groups according to when they are applied, i.e., pre-processing, in-processing and post-processing. 

Pre-processing methods aim to reduce the discrimination and bias in the training data so as to improve the fairness of the trained model. Among the many pre-processing methods~\cite{kamiran2012data, feldman2015certifying, calmon2017optimized, zemel2013learning}, we focus on the following two representatives in this work. 
\begin{itemize}[leftmargin=*]
\item Reweighing (RW)~\cite{kamiran2012data} works by assigning different weights to training samples in order to reduce the effect of data biases. In particular, lower weights are assigned to favored inputs which have a higher chance of being predicted with the favorable label and higher weights are assigned to deprived inputs.
\item Disparate Impact Remover (DIR)~\cite{feldman2015certifying} is based on the disparate impact metric which compares the proportion of individuals that are predicted with the favorable label for an unprivileged group and a privileged group. It modifies the values of the non-protected attribute to remove the bias from the training dataset. 
\end{itemize}


In-processing methods modify the model in different ways to mitigate the bias in the model predictions~\cite{celis2019classification, zhang2018mitigating, kamishima2012fairness, agarwal2018reductions, agarwal2019fair, kearns2018preventing}. We focus on the following representative in-processing methods in this work.

\begin{itemize}[leftmargin=*]
    \item Classification with fairness constraints (META)~\cite{celis2019classification} develops a meta-algorithm which captures the desired metrics of group fairness (e.g., GDS), using convex fairness constraints (with strong theoretical guarantees) and then using the constraints as an additional loss function for training the neural network. 
    
    \item Adversarial debiasing (AD)~\cite{zhang2018mitigating} modifies the original model by including backward feedback for predicting the protected attribute. It maximizes the predictors' ability for classification while minimizing the adversary’s ability to predict the protected attribute to mitigate the bias.
    
    \item Prejudice remover regularizer (PR)~\cite{kamishima2012fairness} focuses on the indirect prejudge. It uses regularizers to compute and restrict the effect of the protected attributes.
    
    \item Exponential gradient reduction (GR)~\cite{agarwal2018reductions} reduces the fair classification problem to a sequence of cost-sensitive classification problems, whose solutions yield a randomized classifier with the lowest empirical error subject to the desired constraints.
\end{itemize}

Post-processing methods modify the prediction results instead of the inputs or the model. We consider three representative processing algorithms in this work.
\begin{itemize}[leftmargin=*]
    \item Equalized Odds (EO)~\cite{hardt2016equality} solves a linear program to find probabilities with which to change the output labels, so as to optimize equalized odds on protected attributes.
    \item Calibrated Equalized Odds (CEO)~\cite{pleiss2017fairness} optimizes over calibrated classifier score outputs to find probabilities with which to change output labels with an equalized odds objective.
    \item Reject Option Classification (RO)~\cite{kamiran2012decision} assigns favorable labels to unprivileged instances and unfavorable labels to privileged instances around the decision boundary with the highest uncertainty. 
\end{itemize}


\section{An Empirical Study} \label{sec:empirical}
In this section, we present an empirical study which aims to compare the performance of different fairness improving methods on different models, different protected attributes or attribute combinations. 

\subsection{Experimental Setup}
\noindent \emph{Datasets} Our experiments are based on 4 models trained with the following benchmark datasets: Census Income~\cite{census1996dataset},  German Credit~\cite{credit1994dataset}, Bank Marketing~\cite{moro2014data} and COMPAS~\cite{compas2016data}. 
These datasets have been used as the evaluation subjects in multiple previous studies~\cite{zhang2020white,galhotra2017fairness,dixon2018measuring,ruoss2020learning,ma2020metamorphic, zhang2021ignorance}.  

\begin{itemize}[leftmargin=*]
\item{Adult Income: 
The prediction task of this dataset is to determine whether the income of an adult is above \$50,000 annually. The dataset contains more than 30,000 samples. The attributes $gender$, $race$ are protected attributes. }
\item{German Credit: 
This is a small dataset with 600 samples. The task is to assess an individual’s credit based on personal and financial records. The attributes $gender$ and $age$ are protected attributes.}
\item{Bank Marketing}:
The dataset contains more than 45,000 samples and is used to train models for predicting whether the client would subscribe a term deposit. Its only sensitive attribute is $age$.
\item{COMPAS: 
The COMPAS Recidivism dataset contains more than 7,000 samples and is used to predict whether the recidivism risk score for an individual is high. The attributes $gender$, $race$ are protected attributes.}
\end{itemize}

In our experiment, we define privileged and unprivileged groups based on the default setting in~\cite{bellamy2019ai}. The details of the privileged group definitions and favorable class are summarised at Table~\ref{tab:privileged}. Altogether, we have a total of 10 model-attribute combinations. Our implementation of the 9 fairness improving methods is based on the AIF360 implementation~\cite{bellamy2019ai}. Each implementation is manually reviewed and tested through standard practice. \\

\begin{table}[t]
\caption{Dataset Privileged Groups Definition}
\label{tab:privileged}
\resizebox{\linewidth}{!}{
\begin{tabular}{lllll}
\toprule
\textbf{Dataset}                        & \textbf{protected Attribute} & \textbf{Privileged Group}    & \textbf{Favorable Class}                       \\
\midrule
\multirow{2}{*}{Adult Income}  & gender            & gender=Male         & \multirow{2}{*}{income\textgreater{}50K}   \\
                               & race              & race=Caucasian      &                                          \\
\midrule                               
\multirow{2}{*}{German Credit} & gender            & gender=Male         & \multirow{2}{*}{good credit}        \\
                               & age               & age\textgreater{}30 &                                          \\
\midrule                               
Bank Marketing & age & age\textgreater{}30 & Yes  \\  
\midrule
\multirow{2}{*}{COMPAS}        & gender            & gender=Female       & \multirow{2}{*}{no recidivism}       \\
                               & race              & race=Caucasian      &                                          \\                                   
\bottomrule
\end{tabular}}
\end{table}

\noindent \emph{Model Training} 
Our models are feed-forward neural networks, which are shown to be highly accurate and efficient in these real-world classification problems~\cite{jain2000statistical, zhang2000neural, abiodun2018state}. All these neural networks contain five hidden layers, each of which contains 64, 32, 16, 8 and 4 units. The output layer contains 2 (number of predict classes) units. For each dataset, we split the data into 70\% training data and 30\% test data. All experiments are conducted on a server running Ubuntu 1804 operating system with 1 Intel Core 3.10GHz CPU, 32GB memory and 2 NVIDIA GV102 GPU. To mitigate the effect of randomness, whenever relevant, we set the same random seed for each test. The trained models reach standard state-of-the-art accuracy. The trained results including the corresponding fairness scores are shown in Table~\ref{tab:NN}. Note that SPD is the probability difference between the unprivileged and privileged groups which is defined on a single protected attribute and thus it is irrelevant if multiple protected attributes are considered simultaneously. 

\begin{table}[t]
\caption{Neural Networks in Experiments}
\label{tab:NN}
\resizebox{\linewidth}{!}{
\begin{tabular}{llllll}
\toprule
\textbf{Dataset} & \textbf{Protected Attribute} & \textbf{SPD} & \textbf{GDS} & \textbf{CDS} & \textbf{Accuracy} \\
\midrule
\multirow{3}{*}{Adult Income} & gender & 0.249 & 0.249 & 0.103 & \multirow{3}{*}{81.7\%} \\
 & race & 0.119 & 0.119 & 0.117 &  \\
 & gender+race & - & 0.306 & 0.179 &  \\
\midrule
\multirow{3}{*}{German Credit} & gender & 0.031 & 0.031 & 0.078 & \multirow{3}{*}{63.3\%} \\
 & age & 0.095 & 0.095 & 0.15 &  \\
 & gender+age & - & 0.133 & 0.172 &  \\
\midrule
Bank & age & 0.047 & 0.047 & 0.014 & 90.0\% \\
\midrule
\multirow{3}{*}{COMPAS} & gender & 0.227 & 0.227 & 0.076 & \multirow{3}{*}{72.7\%} \\
 & race & 0.151 & 0.151 & 0.028 &  \\
 & gender+race & - & 0.301 & 0.083 & \\
 \bottomrule
\end{tabular}}
\end{table}

\subsection{Evaluation Results} \label{sec:empirical_evaluation}
In the following, we present the results of the empirical study, which aims to answer the following research questions.\\ 

\noindent \emph{RQ1: Do the fairness improving methods always improve group fairness?} 
To answer the question, we systematically apply all fairness improving methods on all the model-attribute combinations and measure the effectiveness of the fairness improving methods. We measure the group fairness improvement as follows. SPD is adopted if a single protected attribute is relevant and GDS is adopted if multiple protected attributes are considered at the same time. Note that GDS is the same as SPD with respect to a single protected attribute.

\begin{figure*}[t]
\centering
\includegraphics[width=0.8\linewidth]{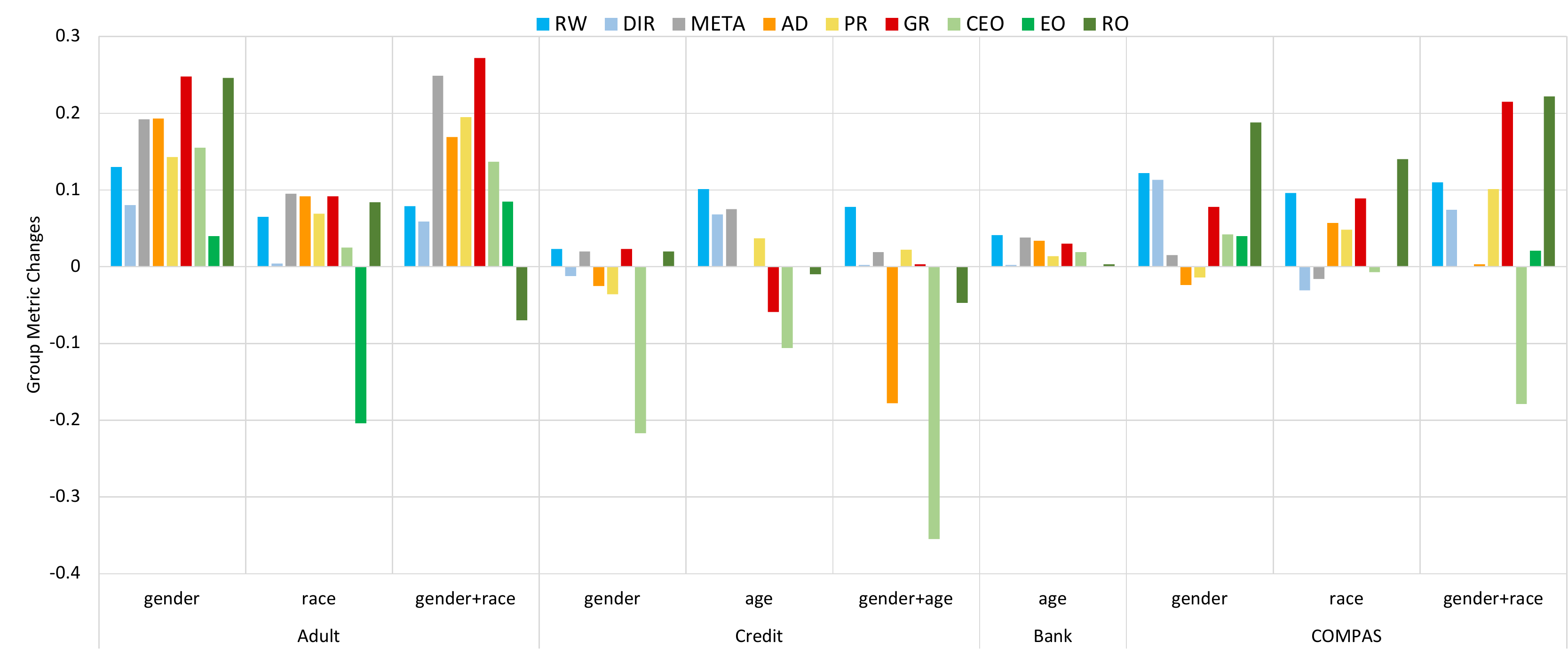}
\centering
\caption{Group Fairness Improvement of Models with respect to Different Protected Attributes}
\label{fig:group_metrics}
\end{figure*}

\begin{figure*}[t]
\centering
\includegraphics[width=0.8\linewidth]{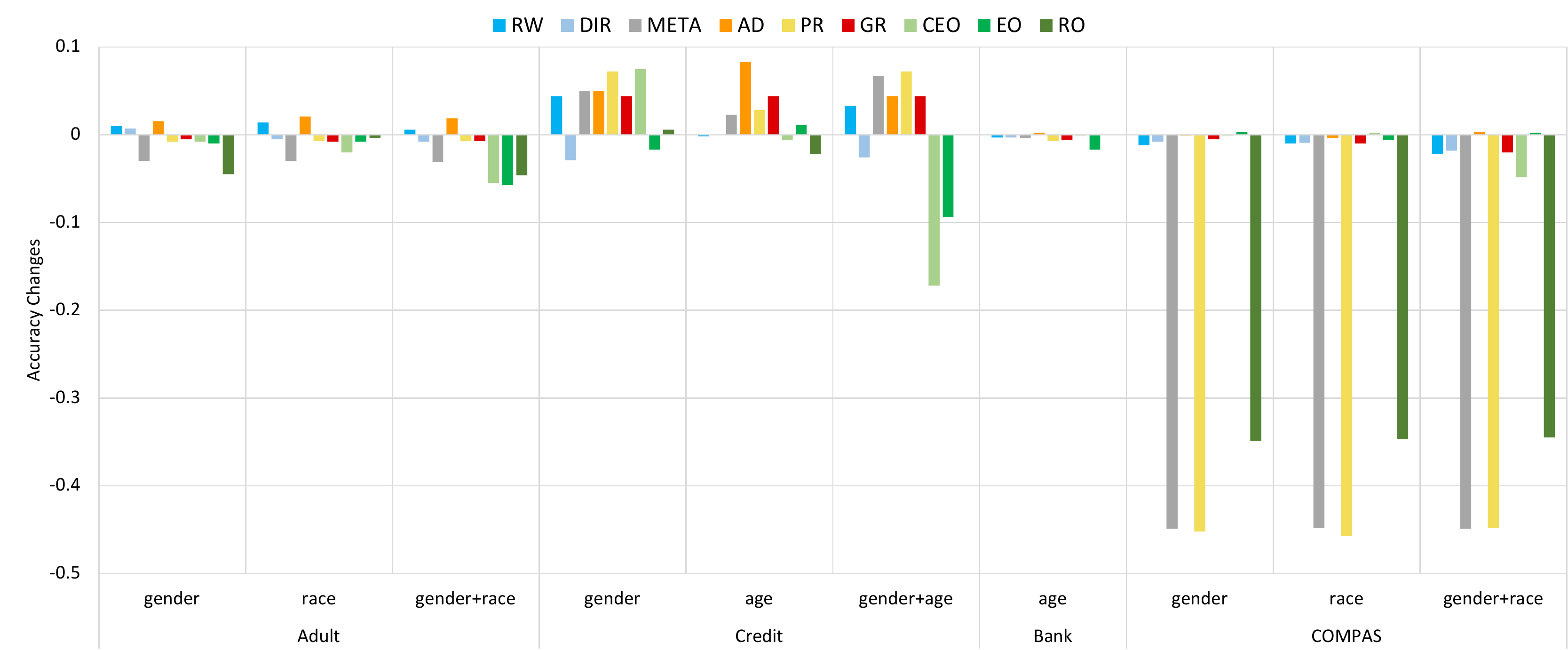}
\centering
\caption{Accuracy Changes of Models with respect to Different Protected Attributes After Processing}
\label{fig:accuracy}
\end{figure*}



The results are shown in Figure~\ref{fig:group_metrics}, where there is one bar for each model-attribute combination and for each fairness improving method, i.e., a total of 9 bars for each model-attribute combination (e.g., \emph{Adult-gender}) and 90 bars in total. A positive value means improved fairness and a negative value means worsened fairness. This bar is shown in 9 different colors for the nine different methods. 

First of all, to our surprise, the fairness improving methods are not always helpful in terms of improving fairness. As shown in Figure~\ref{fig:group_metrics}, while many methods have a positive effect in many cases, \emph{there are many instances where applying fairness improving method results in worsened fairness, sometimes quite significantly.} This is shown as the colorful bar before the zero line, which accounts for a total of 18 cases (i.e., 20\%). Most of those cases are for in-processing and post-processing methods.

Furthermore, the performance of the methods varies significantly across different models and protected attributes. Table~\ref{tab:best_group} shows a summary on which method achieves the most fairness improvement for each model-attribute combination and it can be observed that different winners are there for different model-attribute combinations. Further analysis shows the performance of the fairness improving methods vary across many dimensions. First, the performance of the same method varies significantly on different models. For instance, while the post-processing method CEO works effectively for the neural network trained on Adult Income dataset, it is ineffective for the model trained on German Credit dataset. 
Secondly, the performance of the methods varies across different attributes in the same model. For instance, the post-processing method EO improves the group fairness with respect to \emph{gender} attribute effectively but leads to worse group fairness with respect to \emph{race} attribute for the neural network trained on Adult Income dataset.  

\begin{table}[t]
\caption{Best Method for Group Fairness Improvement}
\label{tab:best_group}
\resizebox{\linewidth}{!}{
\begin{tabular}{llll}
\toprule
\textbf{Dataset} & \textbf{Protected Attribute} & \textbf{Group Fairness} & \textbf{Absolute Change} \\
\midrule
\multirow{3}{*}{Adult Income} & gender & GR & 0.248 \\
 & race & META & 0.095 \\
 & gender, race & GR & 0.272 \\
\midrule
\multirow{3}{*}{German Credit} & gender & RW & 0.023 \\
 & age & RW & 0.101 \\
 & gender, age & RW & 0.078 \\
\midrule
Bank & age & RW & 0.041 \\
\midrule
\multirow{3}{*}{COMPAS} & gender & RO & 0.188 \\
 & race & RO & 0.14 \\
 & gender, race & RO & 0.222 \\
 \bottomrule
\end{tabular}}
\end{table}

Moreover, even the processing methods in the same category behave differently on the same model-attribute combination. In terms of in-processing methods, RW is much more effective than DIR. All models' group fairness can be improved by RW, whereas DIR is ineffective with respect to \emph{Credit-gender} and \emph{COMPAS-race}. For in-processing methods, GR is most effective in improving group fairness for all model-attribute combinations except \emph{Credit-age}. The performance of Post-processing methods varies significantly. For example, the post-processing method RO is much more effective in improving the group fairness for the neural network trained on COMPAS dataset than CEO and EO.

There are some conjectures on why fairness improvement approaches may have different effects on different models and different model-attribute combinations. The main reason is that these methods improve fairness based on certain metrics which may be subtly different from common notions of fairness such as SPG, GDS and CDS. For instance, CEO focuses on reducing False Positive Rate difference in particular, which sometimes translates to fairness measured using SPG/GDS/CDS (as for the Adult Income dataset) and sometimes not. For the different performances on different model-attribute combinations, there may be two reasons. The first is that the discrimination against different attributes in the model may be very different (see in Table~\ref{tab:NN} and observed in~\cite{biswas2020machine}). The second possible reason is that the reasons of the discrimination against different attributes may be different, e.g., biased training data or biased models.

\begin{tcolorbox}[fonttitle = \bfseries]
  \textbf{Answer to RQ1: Existing fairness improving methods are not always effective in improving group fairness and thus they must be applied with caution.} 
\end{tcolorbox}

\noindent \emph{RQ2: What is the cost on accuracy when applying existing fairness improving methods?}
The results are shown in Figure~\ref{fig:accuracy}, where there is similarly one bar for each model-attribute combination and for each fairness improving method. A positive value indicates an increased accuracy and a negative value indicates a decreased accuracy. 

First of all, we observe that some of the fairness improving methods may indeed incur a significant loss of accuracy. This is most observable on META, PR, CEO, EO and RO. Especially for the neural network trained on the COMPAS dataset, the accuracy drops more than 40\% after applying META, PR or RO. 
The average loss of accuracy is around 13\% after processing by META and 12\% after processing by RO. To our surprise, some of the fairness improving methods result in improved accuracy in some cases. This is most observable in some in-processing methods. Especially for the neural network trained on the German Credit dataset, the accuracy increases after applying all four in-processing methods. It should be noted however most of these in-processing methods have a less or harmful effect in terms of group fairness improvement in these cases. For example, while the accuracy increases by 4\% after applying GR on \emph{Credit-age}, the SPD fairness score worsens by 6\%. 

The accuracy reduction varies across not only different model-attribute combinations, but also different methods across different categories. 
Compared fairness improving methods from different categories, the pre-processing methods have an overall mild impact on the model accuracy. In terms of the most effective pre-processing method RW, it is effective on group fairness improvement with respect to all model-attribute combinations and scarifies little accuracy. 
In terms of the most effective in-processing method GR, it is effective on group fairness improvement with respect to all model-attribute combinations except \emph{Credit-age} (although sometimes with minimal fairness improvement). Among them, 7 neural networks get lower accuracy after processing. But the accuracy drops less than 1\% in average. 
In terms of the post-processing method RO, it is effective on group fairness improvement with respect to 7 model-attribute but 5 neural networks get lower accuracy after processing. Especially for the neural network trained on COMPAS dataset, the accuracy drops more than 30\%, which is unacceptable. 
\begin{tcolorbox}[fonttitle = \bfseries]
  \textbf{Answer to RQ2: Existing fairness improving methods may incur a significant loss in accuracy.} 
\end{tcolorbox}

\begin{figure*}[t]
\centering
\includegraphics[width=\linewidth]{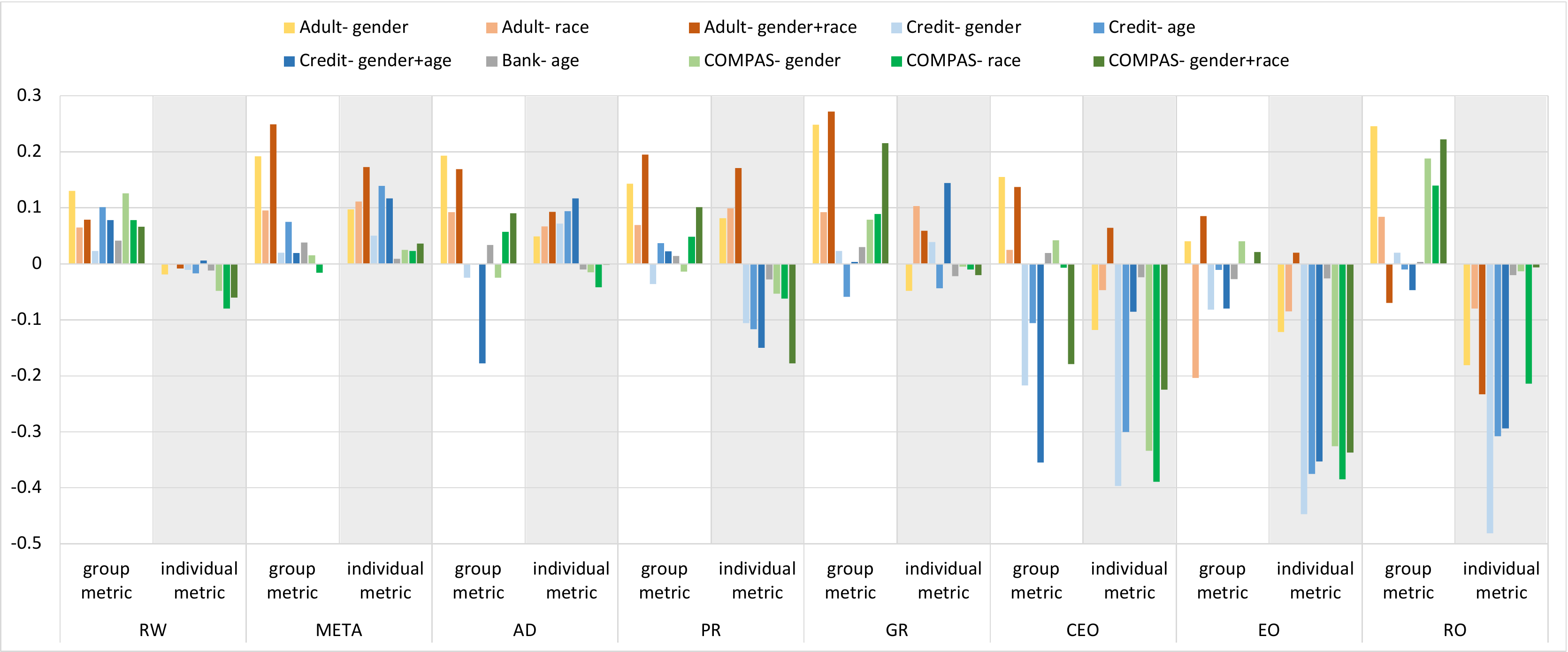}
\centering
\caption{Comparison between group fairness improvement and individual discrimination reduction}
\label{fig:group_individual}
\end{figure*}

\noindent \emph{RQ3: Do the fairness improving methods perform differently for improving group fairness and reducing individual discrimination?} 
Almost all existing fairness improving methods focus on group fairness (whilst some fairness testing approaches focus on individual discrimination for some reason). Thus we are curious about whether the existing fairness improving methods can reduce individual discrimination as well. To answer this question, we compare the CDS change against the group fairness metric change achieved by the same method. The idea is to check whether the changes are consistent, i.e., whether an improvement in group fairness leads to a reduction in individual discrimination and vice versa. Note that, by the default setting in~\cite{bellamy2019ai}, the DIR pre-processing method removes all protected attributes, which makes individual discrimination irrelevant, and thus is not considered in this experiment. 

The results are shown in Figure~\ref{fig:group_individual}, where the CDS change is placed next to the fairness metric change for each fairness improving method. First of all, the group fairness improvement and individual discrimination reduction are inconsistent. A method improving the group fairness effectively might have none or even harmful effect on individual fairness. This is most observable on RW and RO. The pre-processing method RW is effective on group fairness improvement for all models but lead to more individual discrimination for 8 model-attribute combinations. After applying the post-processing method RO, the individual discrimination worsens for all model-attribute combinations.

Furthermore, only the in-processing methods consistently reduce individual discrimination. In terms of META method, it increases the group fairness and reduces the individual discrimination at the same time for 8 model-attribute combinations. The method AD reduces the individual discrimination with respect to all protected attributes in Adult Income dataset and German Credit dataset. Especially for the neural network trained on Adult Income dataset, all in-processing methods improve the individual fairness effectively.
By contrary, all post-processing methods have harmful effect on individual discrimination. on average, the CDS worsens by around 19\% after applying CEO, worsens by 24\% after applying EO and worsens by more than 18\% with RO.

\begin{tcolorbox}[fonttitle = \bfseries]
  \textbf{Answer to RQ3: Existing methods are less effective in reducing individual discrimination.} 
\end{tcolorbox}

\section{An Adaptive Approach} \label{sec:adaptive_approach}
Our empirical study shows that the performance of fairness improving methods varies significantly across different models, i.e., sometimes resulting in worsened fairness and/or reduced accuracy. We thus need a systematic way of choosing the right method. Our proposal is an adaptive approach based on causality analysis. Intuitively, causality analysis measures the ``responsibility'' of each neuron and input attributes towards the unfairness, and depending on whether the most responsible neurons are in the hidden layers or at the input layer, as well as whether a small number of them are significantly more responsible than the rest. Then we choose the fairness improving method accordingly. In the following, we present the details of our approach.

\subsection{Causality Analysis}
Causality analysis aims to identify the presence of causal relationships among events. Furthermore, it can be used to quantify the causal influence of an event on another event. To conduct causality analysis on neural networks, we first adopt the approach in~\cite{chattopadhyay2019neural,causality2022}, and treat neural networks as Structured Causal Models (SCM). Formally, 

\begin{definition}[Structure Causal Model]
A Structure Causal Model consists of a set of endogenous variables $X$ and a set of exogenous variables $U$ connected by a set of functions $F$ that determine the values of the variables in $X$ based on the values of the variables in $U$. The neural network corresponding SCM can be represented as a 4-tuple Model $M(X, U, F, P_U)$, where $P_U$ is the probability of distribution over $U$. 
\hfill \qed
\end{definition}
For the neural network, the endogenous variables $V$ are observed variables, e.g., attributes or neurons. The exogenous variables are the unobserved random variables, e.g., noise, and $P_U$ is the possible distribution of the exogenous variables. Trivially, an SCM can be represented by a directed graphical model $G = (X, E)$, where $E$ is the causal mechanism. 


Based on SCM, the causal effect of a certain event can be computed as the difference between potential outcomes under different treatments. In this work, we adopt the Average Causal Effect (ACE) as the measurement of the causal effect~\cite{chattopadhyay2019neural, causality2022}\footnote{There are alternative ones such as the gradient of causal attribution~\cite{peters2017elements} which work slightly differently.}. The formal definitions of ACE are shown below (where it is assumed that the input endogenous variables are not correlated to each other). 

\begin{definition}[Average Causal Effect]
The ACE of a given endogenous variable $x$ with value $\alpha$ on output $y$ can be measured as: 
\begin{equation}\label{equ:conti_ace}
\begin{aligned}
ACE_{do\left(x=\alpha\right)}^{y}=\mathbb{E}\left[y\mid do\left(x=\alpha\right)\right] - baseline_{x}
\end{aligned}
\end{equation}
where $\mathbb{E}\left[y \mid do\left(x_{i}=\alpha\right)\right]$ represents the interventional expectation which is the expected value of $y$ when the random variable $x$ is set to $\alpha$; and $baseline_{x}$ is the average ACE of $x$ on $y$, i.e., $\mathbb{E}_{x}\left[\mathbb{E}_{y}\left[y \mid d o\left(x=\alpha\right)\right]\right]$\footnote{or alternatively it can be $\mathbb{E}\left[y \mid d o\left(x=\hat{x}\right)\right]$ where $\hat{x}$ is the selected significant point.}. \hfill \qed
\end{definition}


Following the recent work reported in~\cite{causality2022}, we apply ACE to capture the causal influence on model fairness. That is, the $y$ in Equation~\ref{equ:conti_ace} should be a measure of the model unfairness, i.e., SPR, GDS or CDS. For simplicity, we denote it as $y_{fair}$. 

In order to analyze the causal effect on fairness, we analyze two possible causal effects, i.e., the relationship between input attributes to unfairness, and the relationship between the hidden neurons to unfairness. 
In this work, we make use of the average interventional expectation to approximate the ACE of variable $x$ to $y_{fair}$. Formally, $ACE_{do\left(x=\alpha\right)}^{y_{fair}}$ represents the ACE of variable $x$ under value $\alpha$ to the fairness property $y_{fair}$. 
One complication is that each input attribute or neuron has many possible values and we must consider all the possible values in computing the ACE. Our remedy is to consider the average Interventional Expectation (AIE).


\begin{definition}[Average Interventional Expectation]\label{def:inter_expect}
Let $x$ be the given endogenous variable, $y_{fair}$ be the fairness property and $val\_set_{x}$ be a set of values of variable $x$. The average interventional expectation is the mean of expected values of $y_{fair}$ when $x$ is set to be each value $\alpha$: 
\begin{equation}
\begin{aligned}
AIE_{x}^{y_{fair}} = \frac{\sum_{\alpha \in val\_set_{x}}\mathbb{E}[y_{fair} \mid do(x=\alpha)]}{\#(val\_set_{x})} 
\end{aligned}
\end{equation}
\end{definition}

For the input features with categorical values, we intervene the feature with every possible value based on the training dataset. For the hidden neurons with continuous value, intervening it with every possible value might be consuming. We thus intervene the neurons as follows which is adopted in~\cite{chattopadhyay2019neural} as well. That is, we assume the ``intervener'' is equally likely to perturb variable $x$ to any value $\alpha$ within the input range, so that $\alpha \sim U(min_{x}, max_{x})$, where $min_{x}$ and $max_{x}$ are the minimum and maximum input values of $x$. In practice, $min_{x}$ and $max_{x}$ can be obtained by observing the value of the input attribute or neuron given all the training samples and the $val\_set_{x}$ is generated by partitioning the range $[min_{x}, max_{x}]$ uniformly into a fixed number of intervals. Note that if a specific distribution of the interventions is given, it can be used to generate the intervention values instead. 


The details of causality analysis on the hidden neurons are shown in Algorithm~\ref{alg:calsual_neuron}. Given a neural network $N$, a set of inputs $D$ (i.e., the training set), a hidden neuron $n$ and the function for measuring the desired fairness score $fair\_metric$, we systematically measure the AIE with neuron intervention. At line~1 and line~2, we set $min$ to the minimum output of $n$ and $max$ to the maximum output of $n$. Then we generate a set of evenly spaced numbers within the domain of the neuron output $[min, max]$ as $val\_set$ through function $generate\_vals$ at line~3. The input parameter $num\_interval$ decides how many intervals are there. From line~4 to 8, we calculate the AIE with each perturbing value $\alpha$. In each round, we first set $ie$ as an empty set at line~5 and then calculate the fairness score $y_{fair}$ whilst fixing the value of neuron $n$ as $\alpha$. At line~9, we return the mean of all Interventional Expectation as the AIE. 


Algorithm~\ref{alg:causal_attribute} similarly conducts causality analysis on the input attributes. The only difference is that we perform the intervention on the given attribute $f$ at line~4 with all possible values of the attribute. 

\begin{algorithm}[t]
\caption{$CausalityNeuron(N, D, n, fair\_metric)$ where $N$ is the neural network, $D$ is the dataset used to measure causal effect, $n$ is a hidden neuron in $N$ and $fair\_metric$ is the function measuring the fairness score based on the desired fairness metric}
\label{alg:calsual_neuron}
\begin{algorithmic}[1]
\STATE $min \coloneqq$ minimum output of neuron $n$
\STATE $max \coloneqq$ maximum output of neuron $n$
\STATE $val\_set = generate\_vals(min, max, num\_interval)$
\FOR{$\alpha$ in $val\_set$}
\STATE $ie \gets \{\}$
\STATE $y_{fair} = fair\_metric(N, D|do(n=\alpha))$ 
\STATE $ie \gets ie \cup y_{fair}$
\ENDFOR
\RETURN $mean(ie)$
\end{algorithmic} 
\end{algorithm}

\begin{algorithm}[t]
\caption{$CausalityAttribute(N, D, f, fair\_metric)$ where $N$ is the neural network, $D$ is the dataset used to measure causal effect, $f$ is an input attribute and $fair\_metric$ is the function of measuring the fairness score based on the desired fairness metric}
\label{alg:causal_attribute}
\begin{algorithmic}[1]
\STATE $val\_set \coloneqq$ the set of all possible values of attribute $f$
\FOR{$\alpha$ in $val\_set$}
\STATE $ie \gets \{\}$
\STATE $y_{fair} = fair\_metric(N, D|do(f=\alpha))$ 
\STATE $ie \gets ie \cup y_{fair}$
\ENDFOR
\RETURN $mean(ie)$
\end{algorithmic} 
\end{algorithm}

\subsection{Adaptive Fairness Improvement}
Once we compute the causal effect of each neuron and each input attribute on fairness (i.e., responsibility for unfairness), we can then adaptively select the fairness improving methods. For example, if the causal effects of input attributes are relatively high, the unfairness is more likely to be related to the input attributes and likely to be eliminated by pre-processing methods. Similarly, if the interior neurons in the neural network have high causal effects on the fairness property, in-processing methods might be a suitable choice for fairness improvement. 


Formally, to properly compare the casual effects of neurons and input attributes, we first normalize it with respect to a baseline $\overline{y_{fair}}$, which is the fairness score based on the desired fairness metric without any intervention. The baseline $\overline{y_{fair}}$ can be SPD, GDS and CDS as discussed previously. 

We define the causal effects higher than the basic fairness property as high causal effects and vice versa. In other words, only the variable with a causal effect higher than the basic fairness property has the positive causality to unfairness. That is, we only consider those neurons and attributes with a causal effect higher than $\overline{y_{fair}}$ as responsible. 
Next, we measure the proposition of input attributes and neurons that are considered responsible. Given the set of causal effects of all attribute $AIE_{f}$ and the set of causal effects of all neurons $AIE_{n}$, we formally denote the proportion of high causality attributes as $P_f$ and the proportion of high causality neurons as $P_n$ and define them as follows. 
\begin{equation} \label{equ:high_neuron}
\begin{aligned}
P_f = P(AIE > \overline{y_{fair}} ~|~ AIE \in AIE_{f})
\end{aligned}
\end{equation}

\begin{equation} \label{equ:high_attribute}
\begin{aligned}
P_n = P(AIE > \overline{y_{fair}} ~|~ AIE \in AIE_{n})
\end{aligned}
\end{equation}

Furthermore, we measure the distribution of the ``responsibility'' among the input attributes and neurons, since it intuitively has an impact on which fairness improving method should be chosen. For instance, if all input attributes have similar responsibility for unfairness, it is likely hard to pre-process the inputs so as to eliminate the discrimination. Similarly, if all neurons are equally responsible for unfairness, it is complex to improve the fairness by focusing on a few neurons as in~\cite{causality2022}. 
Formally, we use the Coefficient of Variation (CV) to capture the distribution of the causal effects. CV is used to measure the dispersion of data points around the mean. It represents the ratio of the standard deviation to the mean which indicates the degree of variation. In this setting, the larger the CV, the more uneven the distribution of causal effects. We denote the CV of attributes as $CV_{f}$ and the CV of neurons as $CV_{n}$. 

The details of how to select fairness improving methods are shown in algorithm~\ref{alg:selecting_method}. If both the proportion of responsible attributes and responsible neurons is less than a proportion threshold $P\_thres$, few input attributes and neurons are to be blamed for the unfairness. As a result, it is unlikely pre-processing (which focuses on input attributes) or in-processing (which focuses on the hidden neurons) is effective, and thus we choose to apply the post-processing methods. In practice, we set the threshold $P\_thres$ to be 10\%. Otherwise, there are sufficient number of input attributes or neurons that are responsible for unfairness, we then select to apply a pre-processing method if $CV_{f} > CV_{n}$, i.e., the distribution of causal effects is more uneven in the input attributes which means that some of the input attributes are more responsible. Otherwise, an in-processing method is chosen. For pre-processing methods, RW is preferred over DIR, as RW is also feasible to individual fairness metrics. For in-processing methods and post-processing methods, we choose the method with the best improvement and least accuracy cost. \\

\begin{figure}[t]
\centering
\includegraphics[width=\linewidth]{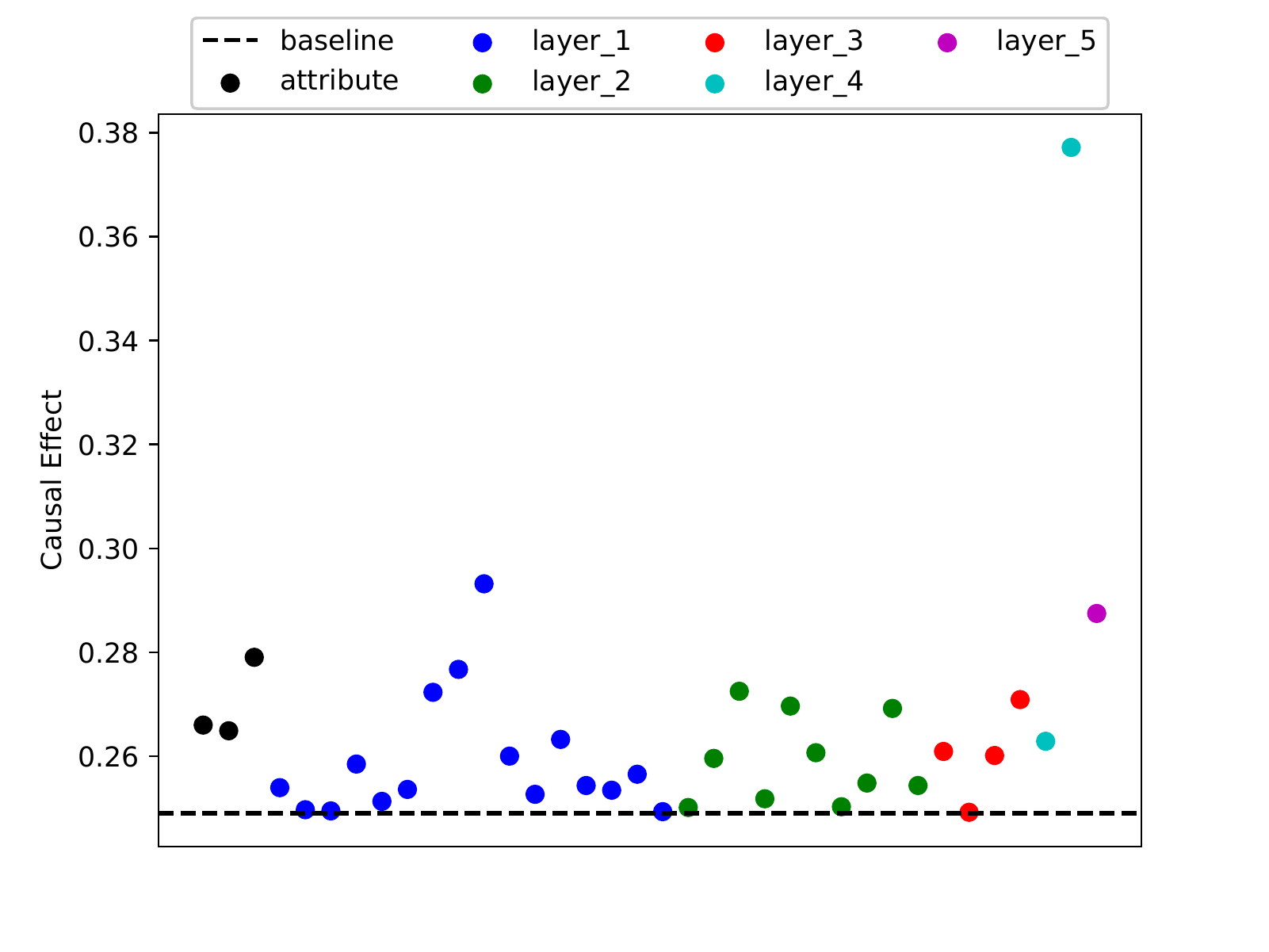}
\centering
\caption{Causality analysis result of \emph{Adult-gender}}
\label{fig:adult_gender_causal_effect}
\end{figure}

\noindent \emph{Example} For the neural network trained on Adult Income dataset, assume that the protected attribute is the ``gender'' attribute. According to the above discussion, we use the group fairness metric SPD to calculate the causal effects of attributes and neurons. The causality analysis result is shown in Figure~\ref{fig:adult_gender_causal_effect}, where each dot represents the AIE of either an input attribute or a hidden neuron. We mark the causal effects of input attributes with black dots and mark the causal effects of hidden neurons in different layers with different colors. The dotted line marks the baseline $\overline{y_{fair}}$ which is 0.249. There are 3 (i.e., 25\%) attributes with causal effects higher than the baseline and 33 (i.e., 26.6\%) neurons with causal effects higher than the baseline. 
As the proportion of responsible input attributes and neurons satisfy the threshold, we then calculate the CV values of those responsible attributes and neurons. The $CV_f$ of these 3 attributes is 0.041 and the $CV_n$ of these 33 neurons is 0.152. Since $CV_n > CV_f$, we choose to apply in-processing methods so as to improve the model's group fairness.
 
\begin{algorithm}[t]
\caption{$AdaptiveImprove(P_n, P_f, CV_n, CV_f)$}
\label{alg:selecting_method}
\begin{algorithmic}[1]
\IF{$P_f \leq P\_thres$ and $P_n \leq P\_thres$}
\RETURN post-processing methods 
\ELSE
\IF{$CV_{f} > CV_{n}$}
\RETURN pre-processing methods
\ELSE
\RETURN in-processing methods
\ENDIF
\ENDIF
\end{algorithmic} 
\end{algorithm}

\section{Implementation and Evaluation} \label{sec:evaluation}
In this section, we evaluate the performance of our adaptive approach systematically to answer multiple research questions. Note that the same datasets, models, and the configuration from Section~\ref{sec:empirical} are used in this section. \\

\noindent \emph{RQ1: How are the ``responsibility'' distributed among the neurons and input attributes} To answer this question, we show the probability of high causal effects and CV of these causal effects for both the hidden neurons and input attributes in Table~\ref{tab:distribution_group} and Table~\ref{tab:distribution_individual}. The first column is the training dataset and the second column shows the corresponding protected attribute(s) in each dataset. Then we show the probability of attributes with high causal effects $P_f$, the probability of neurons with high causal effects $P_n$, CV on highly causal attributes $CV_f$ and CV on highly causal neurons $CV_n$. It can be observed that the distribution of responsibility varies significantly across different model-attribute combinations, which potentially explains why only some fairness improving methods are effective sometimes.

Table~\ref{tab:distribution_group} shows the distribution of high causal effects based on group fairness metrics, e.g., SPD for single protected attributes and GDS for multivariate protected attributes. Based on algorithm~\ref{alg:selecting_method}, the selected processing categories are shown in the last column. 
For all attribute(s) in Adult Income dataset, the probabilities of high causal effects are higher than 10\% and $CV_n$ scores are higher than $CV_f$ scores. So we decide to apply pre-processing methods to this model to improve the group fairness for all attributes. For the neural network trained on German Credit dataset with respect to all attributes, we conclude to apply pre-processing methods. For example, with respect to \emph{age} attribute, both the proportion and the CV of high causal neurons are lower than the two of high causal attributes. Similarly, based on the distribution of high causal effects, we conclude to apply pre-processing to the neural network trained on Bank dataset and the neural network trained on COMPAS dataset with respect to \emph{gender} and \emph{race} attributes. With respect to \emph{gender+race} attribute in COMPAS dataset, as the CV of neurons is higher, we conclude to apply in-processing methods. 

Table~\ref{tab:distribution_individual} show the distribution of high causal effects based on individual fairness metrics, e.g., CDS. The selected processing categories are shown in the last column. Similarly, Algorithm~\ref{alg:selecting_method} decides to apply in-processing methods for all model-attribute combinations, expect \emph{Credit-gender} and \emph{Bank-age}.  We can observe that the proportion of high causal effects of attributes might be 0\% in some cases, e.g., \emph{COMPAS-gender} and \emph{COMPAS-race}, which means no attribute is responsible for individual discrimination. 

Note that, post-processing methods are selected only if both the proportions of responsible neurons/attributes are low, as it often has a significant negative impact on model performance (so that it is impossible to improve fairness through pre-processing or in-processing). In our experiments, however, all the neural networks have sufficiently many responsible neurons/attributes, so no post-processing method is adopted. \\

\begin{table}[t]
\caption{Distribution of high causal effects with Group Fairness}
\label{tab:distribution_group}
\resizebox{\linewidth}{!}{
\begin{tabular}{lllllll}
\toprule
\textbf{Dataset} & \textbf{Protected Attribute} & $P_f$ & $P_n$ & $CV_f$ & $CV_n$ & \textbf{Processing} \\
\midrule
\multirow{3}{*}{Adult Income} & gender & 25.0\% & 26.6\% & 0.041 & 0.152 & in-processing \\
 & race & 16.6\% & 28.2\% & 0.104 & 0.215 & in-processing \\
 & gender+race & 27.3\% & 26.7\% & 0.095 & 0.163 & in-processing \\
\midrule
\multirow{3}{*}{German Credit} & gender & 73.7\% & 46.0\% & 0.339 & 0.323 & pre-processing \\
 & age & 21.1\% & 9.6\% & 0.160 & 0.096 & pre-processing \\
 & gender+age & 77.8\% & 53.2\% & 0.269 & 0.235 & pre-processing \\
\midrule
Bank & age & 33.3\% & 37.9\% & 0.183 & 0.142 & pre-processing \\
\midrule
\multirow{3}{*}{COMPAS} & gender & 63.6\% & 43.5\% & 0.052 & 0.045 & pre-processing \\
 & race & 36.4\% & 19.4\% & 0.056 & 0.034 & pre-processing \\
 & gender+race & 60.0\% & 86.3\% & 0.0018 & 0.002 & in-processing \\
 \bottomrule
\end{tabular}}
\end{table}

\begin{table}[t]
\caption{Distribution of high causal effects with Individual Discrimination}
\label{tab:distribution_individual}
\resizebox{\linewidth}{!}{
\begin{tabular}{lllllll}
\toprule
\textbf{Dataset} & \textbf{Protected Attribute} & $P_f$ & $P_n$ & $CV_f$ & $CV_n$ & \textbf{Processing} \\
\midrule
\multirow{3}{*}{Adult Income} & gender & 75.0\% & 58.8\% & 0.033 & 0.058 & in-processing \\
 & race & 75.0\% & 38.7\% & 0.128 & 0.141 & in-processing \\
 & gender+race & 63.3\% & 46.8\% & 0.091 & 0.105 & in-processing \\
\midrule
\multirow{3}{*}{German Credit} & gender & 94.7\% & 70.2\% & 0.114 & 0.096 & pre-processing \\
 & age & 63.2\% & 29.0\% & 0.041 & 0.053 & in-processing \\
 & gender+age & 83.3\% & 10.3\% & 0..061 & 0.066 & in-processing \\
\midrule
Bank & age & 40.0\% & 50.8\% & 0.076 & 0.047 & pre-processing \\
\midrule
\multirow{3}{*}{COMPAS} & gender & 0\% & 15.3\% & - & 0.026 & in-processing \\
 & race & 0\% & 21.0\% & - & 0.133 & in-processing \\
 & gender+race & 30\% & 39.5\% & 0.075 & 0.1 & in-processing \\

\bottomrule
\end{tabular}}
\end{table}

\begin{figure}[t]
\centering
\includegraphics[width=\linewidth]{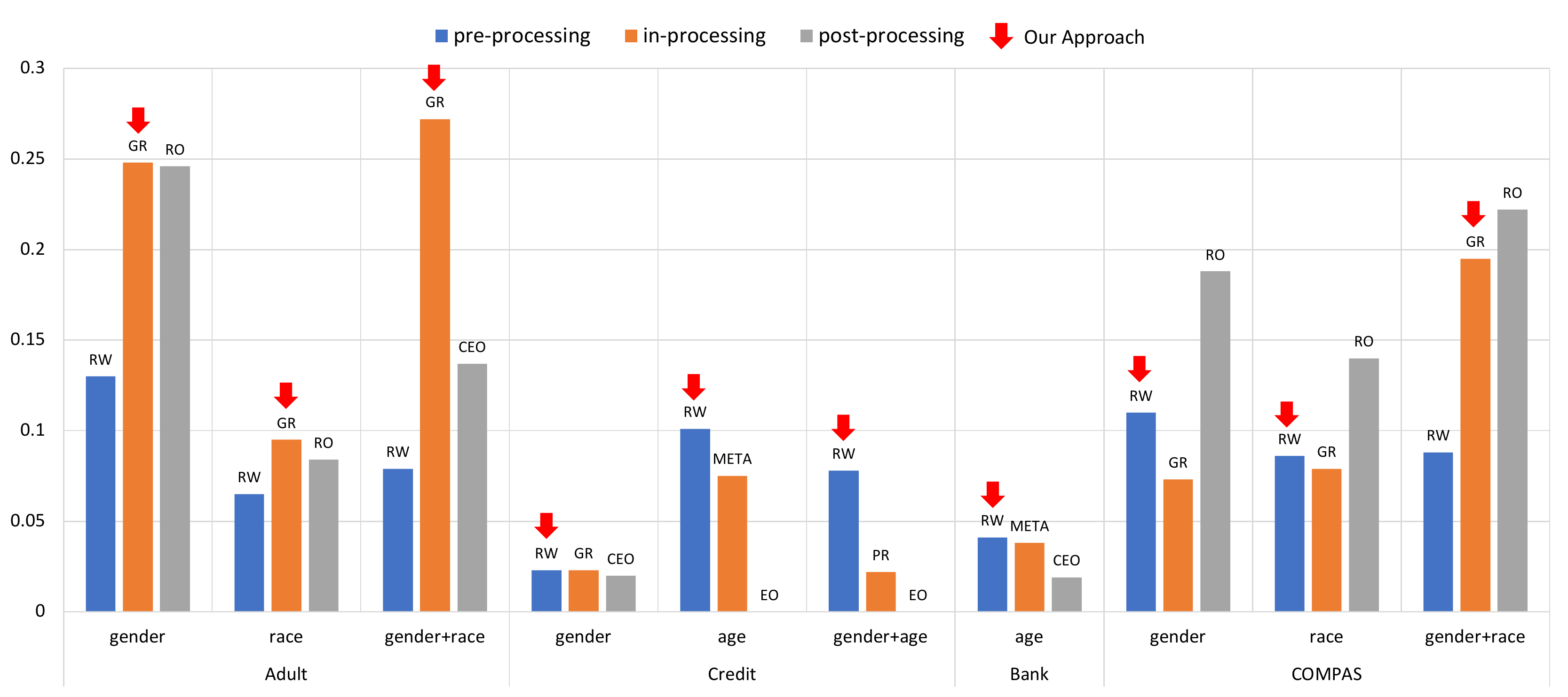}
(a) \emph{Group Fairness Improvement} \\
\centering
\centering
\includegraphics[width=\linewidth]{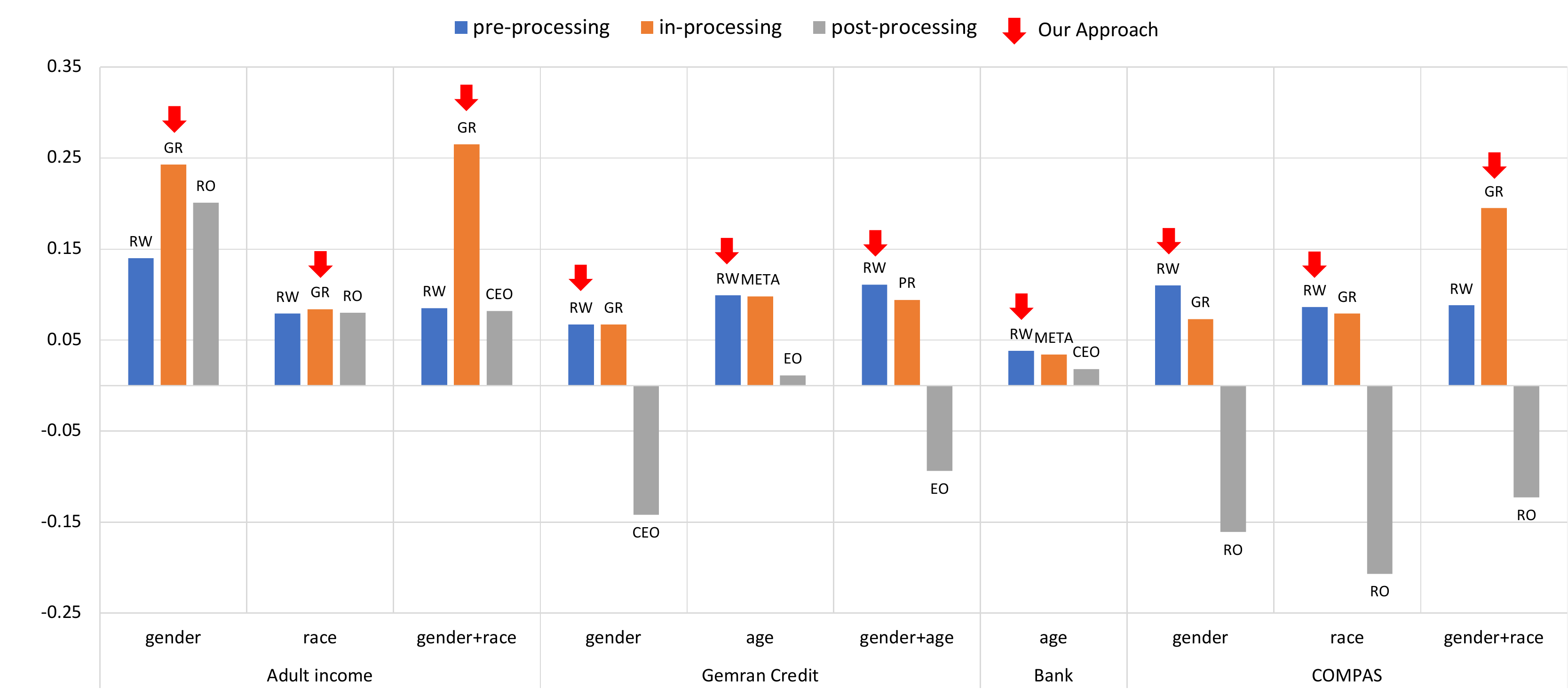}
(b) \emph{Group Fairness Improvement - Accuracy Loss}
\centering
\caption{Our Approach vs SOTA on Group Fairness}
\label{fig:selecting_processing_group}
\end{figure}

\begin{figure}[t]
\centering
\includegraphics[width=\linewidth]{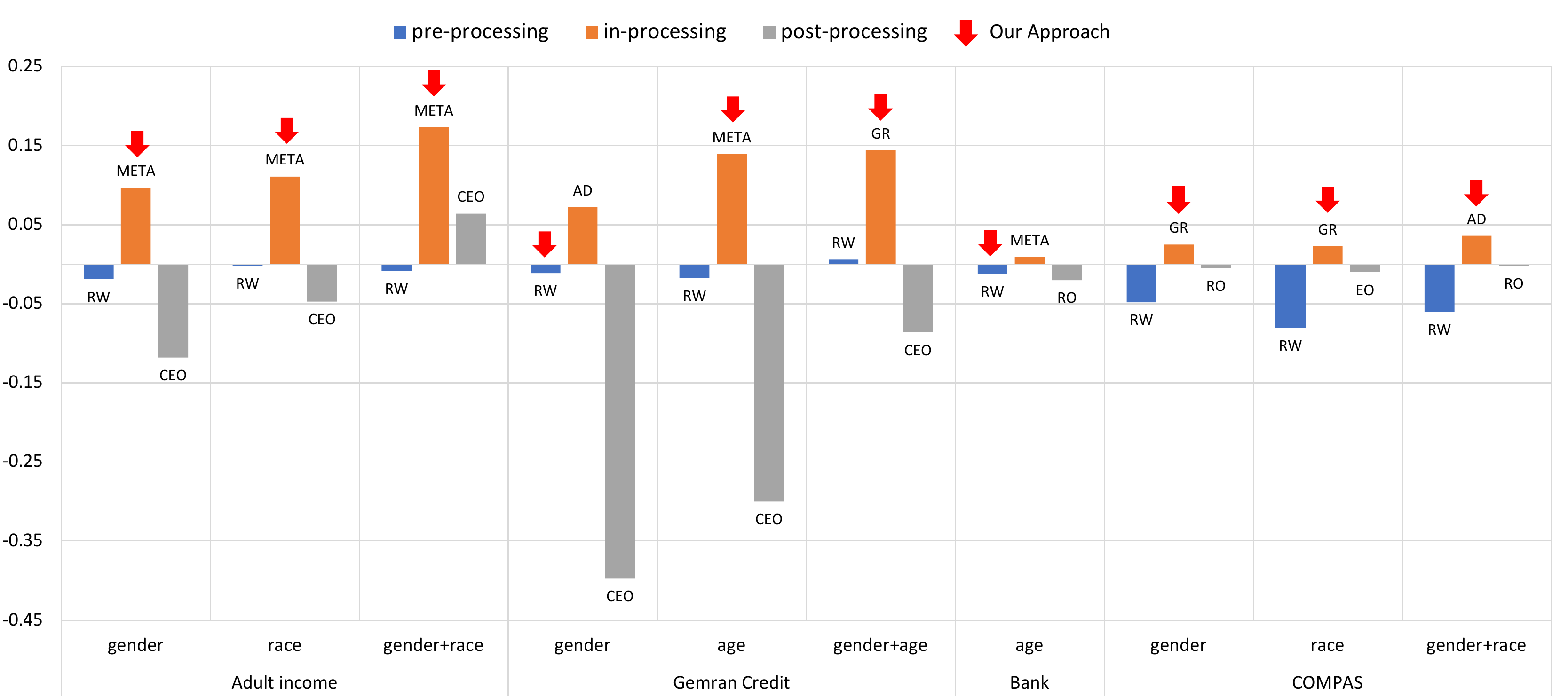}
(a) \emph{Individual Discrimination Reduction} 
\centering
\centering
\includegraphics[width=\linewidth]{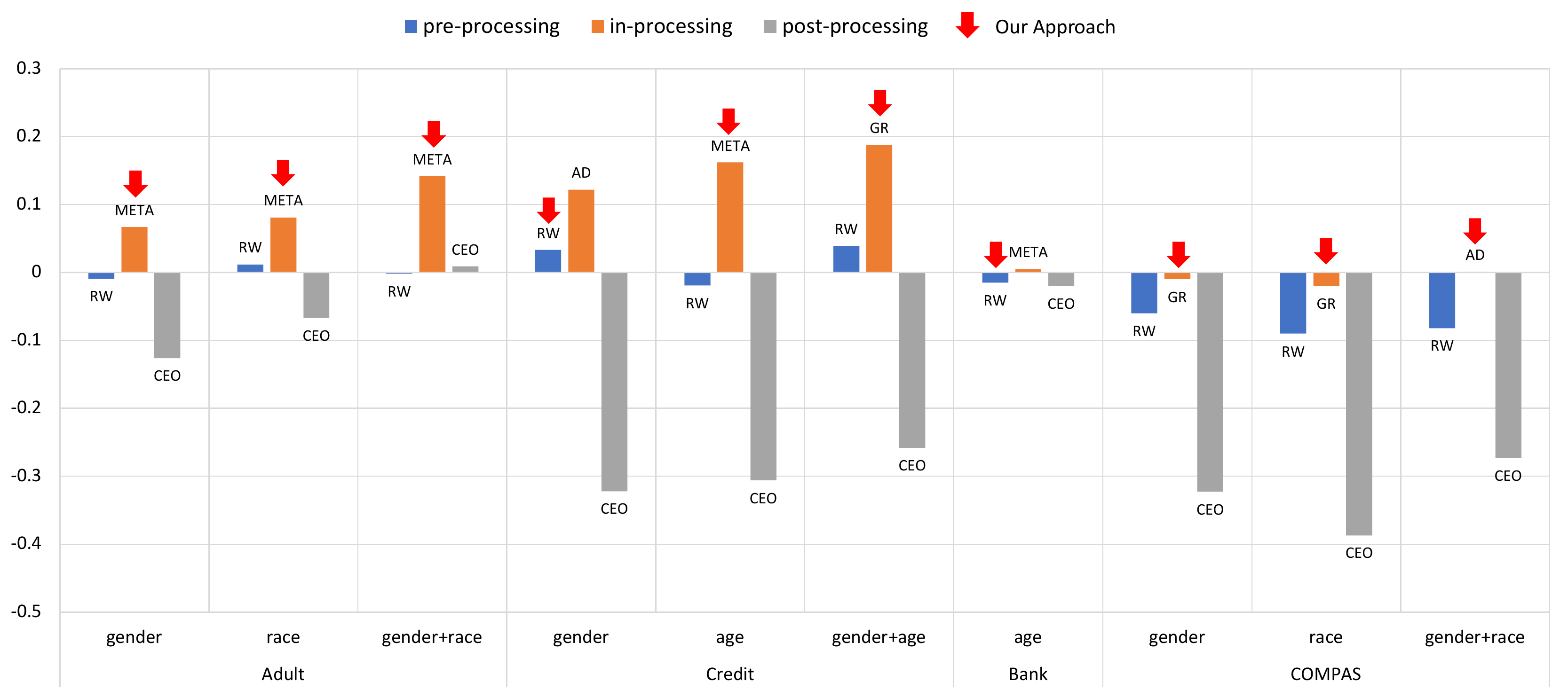}
(b) \emph{Individual Discrimination Reduction - Accuracy Loss}
\centering
\caption{Our Approach vs SOTA on Individual Discrimination}
\label{fig:selecting_processing_individual}
\end{figure}

\noindent \emph{RQ2: Are we always able to identify the best performing fairness improvement method?} To answer this question, we compare our adaptive approach against the best performing pre-processing, in-processing and post-processing method in four ways.
\begin{itemize} [leftmargin=*]
    \item One is the group fairness improvement, which is shown in Figure~\ref{fig:selecting_processing_group}(a).
    \item One is the group fairness improvement minus the accuracy loss, which is shown in Figure~\ref{fig:selecting_processing_group}(b).
    \item One is the individual discrimination reduction, which is shown in Figure~\ref{fig:selecting_processing_individual}(a).
    \item One is the individual discrimination reduction minus the accuracy loss, which is shown in Figure~\ref{fig:selecting_processing_individual}(b).
\end{itemize} 

As shown in Figure~\ref{fig:selecting_processing_group}(a), if we focus on group fairness improvement only, our approach achieves the best performance for 7 out of 10 cases, e.g., \emph{e.g., all attributes in Adult Income dataset, all attributes in German Credit dataset the attribute in Bank dataset}.
Although for the neural network trained on Compas dataset, our adaptive approach does not have the best fairness improvement. If we consider at the same time the accuracy loss, as shown in Figure~\ref{fig:selecting_processing_group}(b), our approach performs the best in all of the cases. Note that while the post-processing method RO often improves the group fairness significantly, the accuracy often drops significantly (e.g., more than 30\% after processing with respect to all protected attributes for the COMPAS dataset, which is clearly unacceptable). In fact, according to our experiments, post-processing should rarely be the choice if we would be maintain high-accuracy.The results shown in Figure~\ref{fig:selecting_processing_group}(b) clearly suggests that our approach is able to improve fairness effectively whilst maintaining a high accuracy. 

In Figure~\ref{fig:selecting_processing_individual}(a), we show the comparison between our approach and the existing approaches in terms of reducing individual discrimination. We can observe that only the in-processing methods can reduce the individual discrimination effectively. In fact, our Adaptive Processing Algorithm~\ref{alg:selecting_method} almost always selects to apply in-processing methods, except for \emph{Credit-gender} and \emph{Bank-age}. After applying the in-processing method RW, the CDS remains almost the same with respect to \emph{Credit-gender} but worsens by around 2\% with respect to \emph{Bank-age}. Taking accuracy loss into account at the same time, we show the individual discrimination reduction minus the accuracy lost in Figure~\ref{fig:selecting_processing_individual}(b). Our approach performs best in 8 out of 10 cases, except for the two cases where RW is selected for \emph{Credit-gender} and \emph{Bank-age}. One potential reason why this is the case is that existing pre-processing methods are not designed for reducing individual discrimination and as a result, even if a small number of input attributes are indeed responsible for the unfairness, existing pre-processing methods such as RW are not able to remove biases in the training set effectively. This calls for research into alternative pre-processing methods for reducing individual discrimination. 

It is worth noting that with our approach, we always (10 out 10) achieve improved group fairness and almost always (9 out 10) achieve reduced individual discrimination, whist achieving a low accuracy loss. \\


\noindent \emph{RQ3: What are the time overhead for causality analysis?} 
The time spent on causality analysis is summarised in Table~\ref{tab:causal_time}. Note that the time is the additional time a user has to spend on applying our method before applying the selected fairness improving method. The time required for causality analysis is always less than 10 minutes.

\begin{table}[t] \small
\caption{Time overhead for causality analysis}
\label{tab:causal_time}
\begin{tabular}{lll}
\toprule
Dataset & Protected Attribute & Time(s) \\
\midrule
\multirow{3}{*}{Adult Income} & gender & 495.26 \\
 & race & 504.72 \\
 & gender, race & 553.42 \\
\midrule
\multirow{3}{*}{German Credit} & gender & 107.79 \\
 & age & 116.56 \\
 & gender, age & 221.72 \\
\midrule
Bank & age & 550.52 \\
\midrule
\multirow{3}{*}{COMPAS} & gender & 106.37 \\
 & race & 152.42 \\
 & gender, race & 162.19 \\
 \bottomrule
\end{tabular}
\end{table}

\section{Threats to Validity} \label{sec:limitation}
\noindent \emph{Limited model structures} We currently support feed-forward neural networks (for tabular data) and convolutional neural networks (for images). It is possible to extend our method to support deep learning architectures such as RNN (for text data) by extending causality analysis to handle feedback loops. We focus on feed-forward NN as existing studies on fairness largely focus on tabular data~\cite{zhang2020white,galhotra2017fairness,dixon2018measuring,ruoss2020learning,ma2020metamorphic, zhang2021ignorance}. \\

\noindent \emph{Limited fairness metrics} We only use SPD and GDS metrics for group fairness and CDS metric for individual fairness. We focus on GPD and GDS as they are the primary focuses of existing works~\cite{angell2018themis, bellamy2019ai, harrison2020empirical, kearns2019empirical, causality2022, zhang2021ignorance}. Given that GPD and GDS are similar with other metrics which consider positive classification rate like Disparate Impact, our method could work for other notions of fairness as well. \\ 

\noindent \emph{Causal effect measurement} ACE is commonly used to evaluate causality~\cite{chattopadhyay2019neural, causality2022}. According to~\cite{chattopadhyay2019neural}, alternative measurements like integrated gradients and gradients of causal effect~\cite{peters2017elements} might suffer from sensitivity and induce causal effects by other input features. \\

\noindent \emph{Distributional shift in the data} Our approach might be affected by distributional shifts in the data. We evaluate the stability of our approach against slight distributional shifts on Adult Income dataset. Firstly, following~\cite{friedler2019comparative}, we randomly split train/test set 10 times and then evaluate whther the method selected by our approach is the best one for each of the 10 test sets. Secondly, following~\cite{udeshi2018automated}, we evaluate our approach using data generated by perturbation. In both conditions, the results confirm that is the case. It shows perhaps that our approach is robust to such levels of distributional shift. 

\section{Related Work} \label{sec:related}
This work is related to research on fairness improving methods, fairness testing and fairness verification methods as well as broadly various studies on fairness. Besides those mentioned in the previous sections, we summarize other related works below. \\

\noindent \emph{Fairness Testing and Verification}
Some existing works attempted to test model discrimination with fairness score measurements. In~\cite{tramer2017fairtest}, Tramer \emph{et al.} propose an unwarranted associations framework to detect unfair, discriminatory or offensive user treatment in data-driven applications. It identifies discrimination according to multiple metrics including the CV score, related ratio and associations between outputs and protected attributes. In~\cite{kleinberg2016inherent}, Kleinberg \emph{et al.} also test multiple discrimination scores and compare different fairness metrics. In~\cite{galhotra2017fairness}, Galhotra \emph{et al.} propose a tool called THEMIS to measure software discrimination. It tests discrimination with two fairness definitions, i.e., group discrimination score and causal discrimination score. It measures these two scores for different software instances with respect to race and gender separately. Their approach generates additional testing samples by selecting random values from the domain for all attributes. In~\cite{adebayo2016iterative}, Adebayo \emph{et al.} try to determine the relative significance of a model's inputs in determining the outcomes and use it to assess the discriminatory extent of the model. In~\cite{ghosh2020justicia}, Ghosh \emph{et al.} verify different fairness measures of the learning process with respect to underlying data distribution. \\

\noindent \emph{Empirical Studies of Fairness} Chakraborty \emph{et al.} empirically research on the effectiveness and efficiency of existing fairness improvement methods based on group fairness metrics~\cite{chakraborty2019software}. Friedler \emph{et al.} work on an empirical study to compare the effects of different fairness improvement methods~\cite{friedler2019comparative}.
In \cite{biswas2020machine}, Biswas \emph{et al.} focus on an empirical evaluation of fairness and mitigation on 8 different real-world machine learning models. They apply 7 mitigation techniques to these models and analyzed the fairness, mitigation results, and
impacts on performance. They also present different trade-off choices of fairness mitigation decisions. Zhang \emph{et al.} discuss how key aspects of machine learning systems, such as attribute set and training data, affect fairness in~\cite{zhang2021ignorance}. Kearns \emph{et al.} test the effectiveness and measure the trade-offs between rich subgroup fairness and accuracy in~\cite{kearns2019empirical}. 
In~\cite{dodge2019explaining}, Dodge \emph{et al.} propose four types of programmatically generated explanations to understand fairness in machine learning systems. 

\section{Conclusion} \label{sec:conclusion}
In this paper, we empirically evaluate 9 fairness improving methods on 4 real world dataset and 90 model-attribute combinations with 3 different fairness metric. Our evaluation shows that existing fairness improving methods are not always effective in improving group fairness and are often not effective in reducing individual discrimination. Meanwhile, we test the trade-off between fairness improvement and accuracy cost. Motivated by the empirical study, we propose a light weight approach to choose the the optimal fairness improving method adaptively based on causality analysis. That is, we identify on the distribution of ``responsible'' attribute and neurons and choose the methods accordingly. Our evaluation shows that our approach is effective in choosing the optimal improvement method.


\section*{Acknowledgement}
This research is supported by the Ministry of Education, Singapore under its Academic Research Fund Tier 3 (Award ID: MOET32020-0004). Any opinions, findings and conclusions or recommendations
expressed in this material are those of the author(s) and do not reflect the views of the Ministry of Education, Singapore.



\balance
\bibliographystyle{ACM-Reference-Format}
\bibliography{ref.bib}

\end{document}